\documentclass[10pt,twocolumn,letterpaper]{article}

\usepackage[dvipsnames]{xcolor}
\usepackage{iccv}
\usepackage{times}
\usepackage{epsfig}
\usepackage{graphicx}
\usepackage{amsmath}
\usepackage{amssymb}
\usepackage[utf8]{inputenc} 
\usepackage[T1]{fontenc}  
\usepackage{color}
\usepackage{booktabs}
\usepackage{amsfonts}       
\usepackage{nicefrac}      
\usepackage{microtype}    
\usepackage[english]{babel}
\usepackage{tabularx}
\usepackage{tabulary}
\usepackage{subfig}
\usepackage{adjustbox}
\usepackage{xspace}
\usepackage{amssymb} 
\usepackage{amsmath} 
\usepackage{amsthm}
\usepackage{float}
\usepackage{graphicx}
\usepackage{tabularx}
\usepackage{wrapfig}
\usepackage{footnote}
\usepackage{comment}
\makesavenoteenv{tabular}
\makesavenoteenv{table}
\usepackage{multirow}
\usepackage{array}
\usepackage[flushleft]{threeparttable}
\usepackage{tikz}
\usepackage{amsmath}
\usepackage[english]{babel}
\usepackage{blindtext}
\usepackage{bbold}
\usepackage{filecontents}
\usepackage{pgfplots}
    \usetikzlibrary{
        matrix,
    }
        \usepgfplotslibrary{colorbrewer}
    \pgfplotsset{
        cycle list/Set1-9,
        cycle multiindex* list={
            mark list*\nextlist
            Set1-9\nextlist
        },
    }
    
    \usepgfplotslibrary{fillbetween}
\usetikzlibrary{positioning}
\usetikzlibrary{backgrounds}
\usetikzlibrary{patterns,arrows}
   \tikzset{
        hatch distance/.store in=\hatchdistance,
        hatch distance=10pt,
        hatch thickness/.store in=\hatchthickness,
        hatch thickness=2pt
    }

    \makeatletter
    \pgfdeclarepatternformonly[\hatchdistance,\hatchthickness]{flexible hatch}
    {\pgfqpoint{0pt}{0pt}}
    {\pgfqpoint{\hatchdistance}{\hatchdistance}}
    {\pgfpoint{\hatchdistance-1pt}{\hatchdistance-1pt}}%
    {
        \pgfsetcolor{\tikz@pattern@color}
        \pgfsetlinewidth{\hatchthickness}
        \pgfpathmoveto{\pgfqpoint{0pt}{0pt}}
        \pgfpathlineto{\pgfqpoint{\hatchdistance}{\hatchdistance}}
        \pgfusepath{stroke}
    }
    \makeatother

\usepackage[acronym]{glossaries}
\usepackage{colortbl}
\usepackage{algorithm}
\usepackage{algorithmic}
\usepackage{float}
\usepackage{multirow}
\usepackage[font=small,labelfont=bf,textfont={}]{caption}
\usepackage{enumitem}
\usepackage{todonotes}

\usepackage[title]{appendix}

\usepackage[pagebackref=true,breaklinks=true,letterpaper=true,colorlinks,bookmarks=false]{hyperref}
\usepackage[capitalize]{cleveref}

\newcommand{\para}[1]{\noindent \textbf{#1}\xspace}
\newcommand{\sys}{\mbox{AMS}\xspace}
\newcommand{\early}{\mbox{One-Time}\xspace}
\newcommand{\frozen}{\mbox{No Customization}\xspace}
\newcommand{\jitcustom}{\mbox{Just-In-Time}\xspace}
\newcommand{\remoteinference}{\mbox{Remote+Tracking}\xspace}

\newcommand{\gstrat}{\mbox{gradient-guided}\xspace}
\newcolumntype{a}{>{\columncolor{lightgray}}c}

\glsdisablehyper
\newacronym{ml}{ML}{Machine Learning}
\newacronym{mdp}{MDP}{Markov Decision Process}
\newacronym{mpc}{MPC}{Model Predictive Controller}
\newacronym{qoi}{QoI}{Quality of Inference}
\newacronym{dnn}{DNN}{Deep Neural Network}

\newacronym{fps}{fps}{frame-per-second}
\newacronym{miou}{mIoU}{mean Intersection-over-Union}

\iccvfinalcopy %

\def\iccvPaperID{X} %

\ificcvfinal\fi

\begin{document}

\title{Real-Time Video Inference on Edge Devices via Adaptive Model Streaming}
\author{Mehrdad Khani, Pouya Hamadanian, Arash Nasr-Esfahany, Mohammad Alizadeh\\
MIT CSAIL\\
{\tt\small \{khani,pouyah,arashne,alizadeh\}@csail.mit.edu}}
\maketitle
\ificcvfinal\thispagestyle{empty}\fi

\begin{abstract}
Real-time video inference on edge devices like mobile phones and drones is challenging due to the high computation cost of Deep Neural Networks. We present Adaptive Model Streaming (AMS), a new approach to improving performance of efficient lightweight models for video inference on edge devices. AMS uses a remote server to continually train and adapt a small model running on the edge device, boosting its performance on the live video using online knowledge distillation from a large, state-of-the-art model.  We discuss the challenges of over-the-network model adaptation for video inference, and present several techniques to reduce communication cost of this approach: avoiding excessive overfitting, updating a small fraction of important model parameters, and adaptive sampling of training frames at edge devices. On the task of video semantic segmentation, our experimental results show 0.4--17.8 percent mean Intersection-over-Union improvement compared to a pre-trained model across several video datasets. Our prototype can perform video segmentation at 30 frames-per-second with 40 milliseconds camera-to-label latency on a Samsung Galaxy S10+ mobile phone, using less than 300 Kbps uplink and downlink bandwidth on the device. 
\end{abstract}
\section{Introduction}
\label{sec:intro}

 \begin{figure}
     \centering
     \includegraphics[trim=1cm 0cm 0 0cm, width=\linewidth]{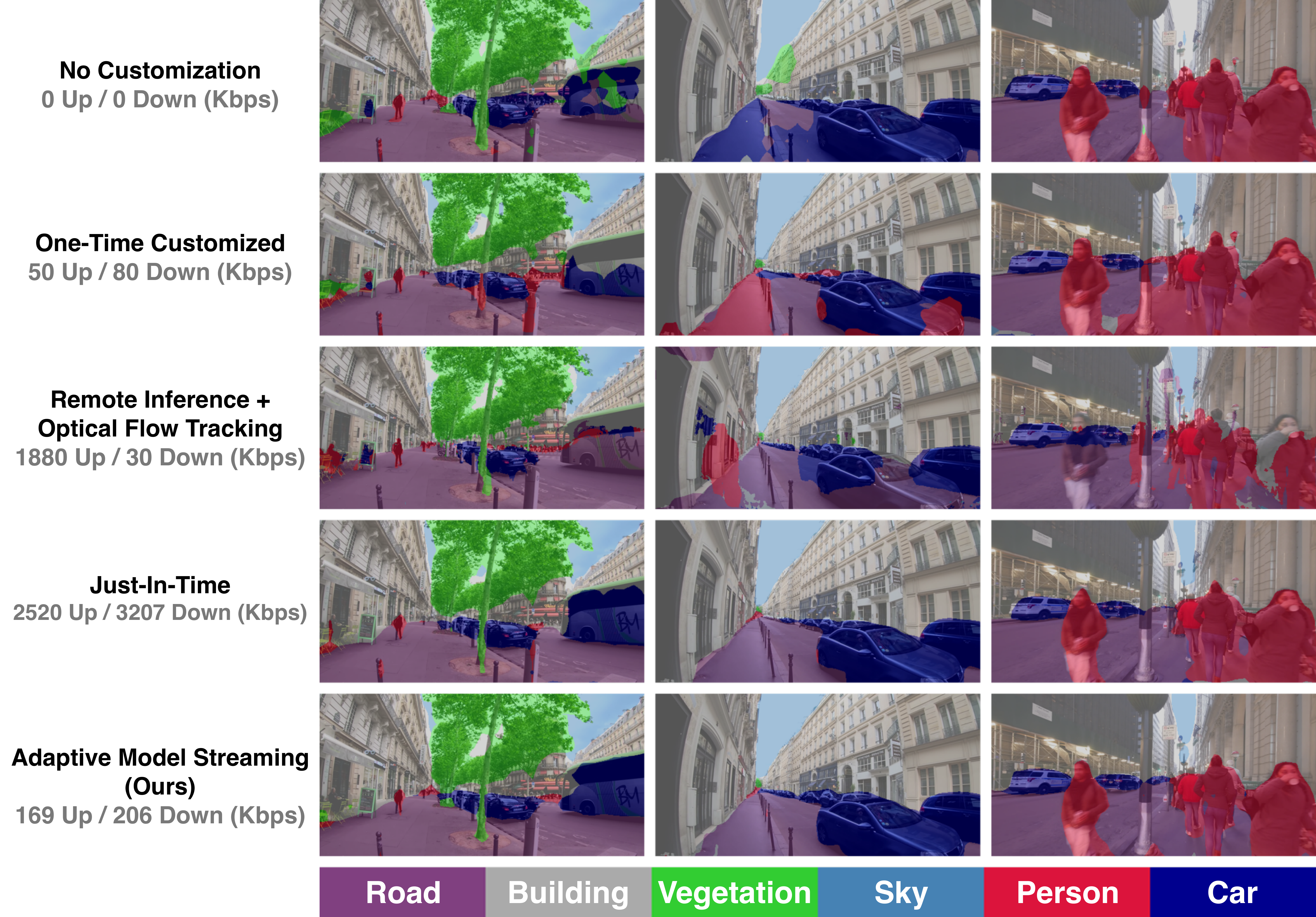}
     \caption{Semantic segmentation results on real-world outdoor videos: rows from top to bottom represent \frozen, \early, \remoteinference, \jitcustom, and \sys. Uplink and downlink bandwidth usage are reported below each variant. \sys provides better accuracy with limited bandwidth and reduces artifacts (e.g., see the car/person detected in error by the no/one-time customized models and remote tracking in the second column).}
     \label{fig:sample_segmentations}
     \vspace{-17pt}
 \end{figure}
 
Real-time video inference is a core component for many applications, such as augmented reality, drone-based sensing, robotic vision, and autonomous driving. These applications use Deep Neural Networks (DNNs) for inference tasks like object detection~\cite{redmon2016you}, semantic segmentation~\cite{Deeplab}, and pose estimation~\cite{cao2017realtime}. However, state-of-the-art DNN models are too expensive to run on low-powered edge devices (e.g., mobile phones, drones,  consumer robots~\cite{deeplab_model_zoo,model_zoo}), and cannot run in real-time even on accelerators such as Coral Edge TPU and NVIDIA Jetson~\cite{coral_benchmark,MohammadSuggestion,Liang2020AIOT}. 

A promising approach to improve inference efficiency is to specialize a lightweight model for a specific video and task. The basic idea is to use distillation~\cite{hinton2015distilling} to transfer knowledge from a large ``teacher'' model to a small ``student'' model. For example, Noscope~\cite{kang2017noscope} trains a student model to detect a few object classes on specific videos offline. Just-In-Time~\cite{mullapudi2019online} extends the idea to live, dynamic videos by training the student model online, specializing it to video frames as they arrive. These approaches provide significant speedups for scenarios that perform inference on powerful machines (e.g., server-class GPUs), but they are impractical for on-device inference at the edge. The offline approach isn't desirable since videos can vary significantly from device to device (e.g., different locations, lighting conditions, etc.), and over time for the same device (e.g., a drone flying over different areas). On the other hand, training the student model online on edge devices is computationally infeasible. 

In this paper we propose {\em Adaptive Model Streaming} (AMS), a new approach to real-time video inference on edge devices that offloads knowledge distillation to a remote server communicating with the edge device over the network. \sys~{\em continually} adapts a small student model running on the edge device to boost its accuracy for the specific video in real time. The edge device periodically sends sample video frames to the remote server, which uses them to fine-tune (a copy of) the edge device's model to mimic a large teacher model, and sends (or ``streams'') the updated student model back to the edge device.

Performing knowledge distillation over the network introduces a new challenge: communication overhead. Prior techniques such as Just-In-Time aggressively overfit the student model to the most recent frames, and therefore must frequently update the model to sustain high accuracy. We show, instead, that training the student model over a suitably chosen horizon of recent frames\,---\,not too small to overfit narrowly, but not too large to surpass the generalization capacity of the model\,---\,can achieve high accuracy with an order of magnitude fewer model updates compared to Just-In-Time training.

Even then, a na\"ive implementation of over-the-network model training would require significant bandwidth. For example, sending a (small) semantic segmentation model such as DeeplabV3 with MobileNetV2~\cite{MobileNet-v2} backbone with $\sim$2 million (float16) parameters every 10 seconds would require over 3 Mbps of bandwidth. We present techniques to reduce both downlink (server to edge) and uplink (edge to server) bandwidth usage for \sys. For the downlink, we develop a coordinate-descent~\cite{coordesc,Nesterov} algorithm to train and send a small fraction of the model parameters in each update. Our method identifies the subset of parameters with the most impact on model accuracy, and is compatible with optimizers like Adam~\cite{Adam} that maintain a state (e.g., gradient moments) across training iterations. For the uplink, we present algorithms that dynamically adjust the frame sampling rate at the edge device based on how quickly scenes change in the video. Taken together, these techniques reduce downlink and uplink bandwidth to only 181--225 Kbps and 57--296 Kbps respectively (across  different videos) for a challenging semantic segmentation task. To put \sys's bandwidth requirement in perspective, it is less than the YouTube recommended bitrate range of 300--700~Kbps to live stream video at the lowest (240p) resolution~\cite{youtuberates}.

We evaluate our approach for real-time semantic segmentation using a lightweight model (DeeplabV3 with MobileNetV2~\cite{MobileNet-v2} backbone). This model runs at 30 frames-per-second with 40~ms camera-to-label latency on a Samsung~Galaxy~S10+ phone~(with Adreno 640 GPU). Our experiments use four datasets with long (10 minutes+) videos spanning a variety of scenarios (e.g., city driving, outdoor scenes, and sporting events). Our results show:
\begin{enumerate}
    \item Compared to pretraining the same lightweight model without video-specific customization, \sys provides a 0.4--17.8\% boost (8.3\% on average) in \gls{miou}, computed relative to the labels from a state-of-the-art DeeplabV3 with Xception65~\cite{Xception65} backbone model. It also improves mIoU by 4.3\% on average (up to 39.1\%) compared to customizing the model once using the first 60 seconds of each video.
    \item
    Compared to a remote inference baseline accompanied by on-device optical flow tracking~\cite{zhu2017deep,beauchemin1995computation}, \sys provides an average improvement of 5.8\% (up to 24.4\%) in \gls{miou}.
    \item \sys requires 15.7$\times$ less downlink bandwidth on average (up to 44.5$\times$) to achieve similar accuracy compared to Just-In-Time~\cite{mullapudi2019online} (with similar reductions in uplink bandwidth).
\end{enumerate}

\Cref{fig:sample_segmentations} shows three visual examples comparing the accuracy of \sys with these baseline approaches.  Our code and video datasets are available online at \href{https://github.com/modelstreaming/ams}{https://github.com/modelstreaming/ams}.

\section{Related Work}\label{sec:related_work}
We described prior work on knowledge distillation for video in \S\ref{sec:intro}. Here, we discuss other related work. 

\para{On-device inference.} Small, mobile-friendly models have been designed both manually~\cite{MobileNet-v2} and using neural architecture search~\cite{zoph2016neural,wu2019fbnet}. Model quantization and weight pruning~\cite{he2018amc,lin2016fixed,blalock2020state,CompressionSurvey} have further been shown to reduce the computational footprint of such models with a small loss in accuracy. Specific to video, some techniques amortize the inference cost by using optical flow methods to skip inference for some frames~\cite{zhu2018distractor,zhu2017deep,hu2017maskrnn}. 
Despite this progress, there remains a large gap in the performance of lightweight models and state-of-the-art solutions~\cite{du2019spinenet,mobilenetv3}. \sys is complementary to on-device optimization techniques and would also benefit from them. 

\para{Remote inference.} Several proposals offload all or part of the computation to a remote machine~\cite{kang2017neurosurgeon,chinchali2018neural,chinchali2019network,olston2017tensorflow,crankshaw2017clipper,zhang2019mark}, but these schemes generally require high network bandwidth, incur high latency, and are susceptible to network outages~\cite{fouladi2018salsify,crankshaw2017clipper}. Proposals like edge computing~\cite{edge1,edge3,edge5} that place the remote machine close to the edge devices lessen these barriers, but do not eliminate them and incur additional infrastructure and maintenance costs. \sys requires much less bandwidth than remote inference, and is less affected by network delay or outages since it performs inference locally on the device.

\para{Online learning.} Our work is also related to online learning~\cite{shalev2012online} algorithms for minimizing dynamic or tracking  regret~\cite{hall2013dynamical, zhang2018adaptive,mokhtari2016online}. Dynamic regret compares the performance of an online learner to a sequence of optimal solutions. In our case, the goal is to track the performance of the best lightweight model at each point in a video. 
Several theoretical works have studied online gradient descent algorithms in this setting with different assumptions about the loss functions~\cite{zinkevich2003online,hazan2007logarithmic}. Other work has focused on the ``experts'' setting~\cite{herbster1998tracking,yang2016tracking,hazan2009efficient}, where the learner maintains multiple models and uses the best of them at each time. Our approach is based on online gradient descent because tracking multiple models per video at a server is expensive. 

Other paradigms for model adaptation include lifelong/continual learning~\cite{lopez2017gradient}, meta-learning~\cite{finn2017model,ravi2017optimization},  federated learning~\cite{mcmahan2017communication}, and unsupervised domain adaptation~\cite{ben2007analysis,kang2019contrastive}. As our work is only tangentially related to these efforts, we defer their discussion to Appendix~\ref{app:related}.

\section{Adaptive Model Streaming (AMS)}\label{sec:framework}
\begin{figure}
    \centering
    {
    \includegraphics[width=\linewidth]{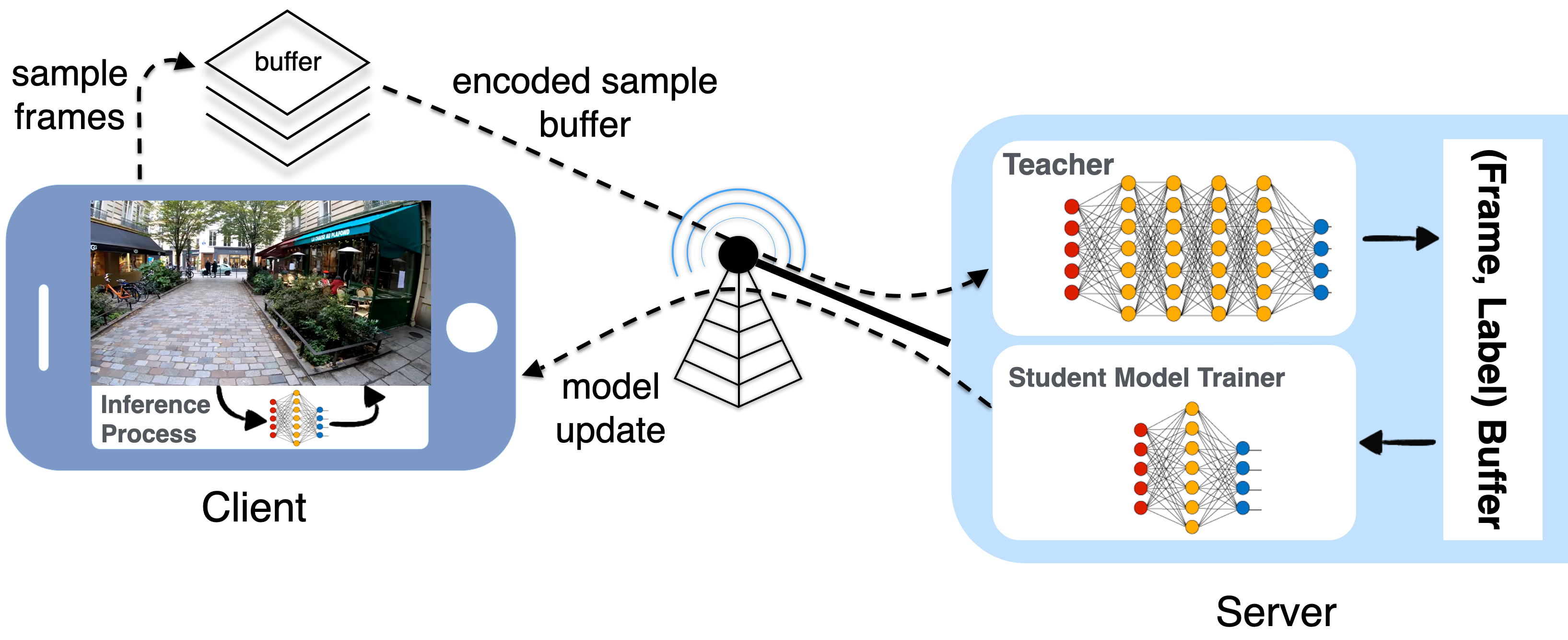}}
    \caption{\sys system overview.}
    \label{fig:ams_overview}
\end{figure}

\cref{fig:ams_overview} provides an overview of AMS. Each edge device buffers sampled video frames for $T_{update}$ seconds, then compresses and sends the buffered frames to the remote server. The server uses these frames to train a copy of the edge device's model using supervised knowledge distillation~\cite{hinton2015distilling}, and sends the model changes to the edge device. For concreteness, we describe our design for semantic segmentation, but the approach is general and can be adapted to other tasks.

\para{Server.} Algorithm~\ref{alg:server_training} shows the server procedure for serving a single edge device (we discuss multiple edge devices in Appendix~\ref{app:server_utilization}). The \sys algorithm at the server runs iteratively on each new batch of frames received from the edge device. It consists of two phases: inference and training. 

\emph{Inference phase:} To train, the server first needs to label the incoming video frames. It obtains these labels using a state-of-the-art segmentation model (like DeeplabV3~\cite{Deeplab} with Xception65~\cite{Xception65} backbone), which serves as the ``teacher'' for supervised knowledge distillation. The server runs the teacher on new frames, and adds the frames,  their timestamps, and labels to a training data buffer $\mathcal{B}$. 

\emph{Training phase:} The server trains the student model to minimize the loss over the sample frames in its buffer from the last $T_{horizon}$ seconds of video. To reduce bandwidth usage, the server selects a small subset (e.g., 5\%) of parameters for each model update, and trains them for $K$ iterations on randomly-sampled mini-batches of frames. %
We discuss how the server chooses the parameters to train in~\S\ref{sec:downlink}.

The server also dynamically adapts the frame sampling rate used by the edge device based on the video characteristics (how fast scenes changes) as described in \S\ref{sec:design_uplink}.

\para{Edge device.} The edge device deploys the new models as soon as they arrive to perform local inference. To switch models without disrupting inference, the edge device maintains an inactive copy of the running model in memory and applies the model update to that copy. Once ready, it swaps the active and inactive models. The edge device also samples frames at the rate specified by the server and sends them to the server every $T_{update}$ seconds.

\begin{algorithm}[t]
\begin{tikzpicture}
\node[draw=none, inner sep=0] (alg) {
\begin{varwidth}{\linewidth}
\footnotesize
\caption{Adaptive Model Streaming Server} 
\label{alg:server_training}
\begin{algorithmic}[1]
\STATE Initialize the student model with pre-trained parameters $\bold{w}_0$
\STATE Send $\bold{w}_0$ and the student model architecture for the edge
\STATE $\mathcal{B}\leftarrow$ Initialize a time-stamped buffer to store (sample frame, teacher prediction) tuples
\FOR{$n \in \{1,2,...\}$}
    \STATE $\mathcal{R}_n \leftarrow$ Set of new sample frames from the edge device
    \FOR {$\bold{x} \in \mathcal{R}_n$ }
        \STATE $\tilde{\bold{y}} \leftarrow$ Use the teacher model to infer the label of $\bold{x}$
        \STATE Add ($\bold{x}, \tilde{\bold{y}})$ to $\mathcal{B}$ with time stamp of receiving $\bold{x}$
    \ENDFOR
    \STATE $\mathcal{I}_n \leftarrow$ Select a subset of model parameter indices
    \FOR {k $\in \{1,2,...,K\}$} 
        \STATE $\bold{S}_k \leftarrow$ Uniformly sample a mini-batch of data points from $\mathcal{B}$ over the last $T_{horizon}$ seconds 
        \STATE Candidate updates $\leftarrow$ Calculate Adam optimizer updates w.r.t the empirical loss on $\bold{S}_k$
        \STATE Apply candidate updates to model parameters indexed by $\mathcal{I}_n$
    \ENDFOR
    \STATE $\tilde{\bold{w}}_n\leftarrow$ New value of model parameters which are indexed by $\mathcal{I}_n$
    \STATE Send ($\tilde{\bold{w}}_n$, $\mathcal{I}_n$) for the edge device
    \STATE Wait for $T_{update}$ seconds 
\ENDFOR
\end{algorithmic}
        \end{varwidth}
            };
    \node[inner sep=0] at (-3.5, .25) (nodeA) {};
    \node[inner sep=0] at (-3.5, 1.9) (nodeB) {};
    \draw[<->, color=black!40, line width=1pt] (nodeA) -- (nodeB) node [midway, sloped,font=\scriptsize, color=black!40, fill=white, inner sep=0] (TextNode) {Infer. Phase};
    
    \node[inner sep=0] at (-3.5, -2.3) (nodeC) {};
    \node[inner sep=0] at (-3.5, 0.2) (nodeD) {};
    \draw[<->, color=black!40, line width=1pt] (nodeC) -- (nodeD) node [midway, sloped,font=\scriptsize, color=black!40, fill=white, inner sep=0] (TextNode) {Training Phase};
    \end{tikzpicture}
    \end{algorithm}

\subsection{Reducing Downlink Bandwidth}
\label{sec:downlink}

The downlink (server-to-edge) bandwidth depends on (i) how frequently we update the student model, (ii) the cost of each model update. We discuss each in turn.

\subsubsection{How Frequently to Train?}
\label{sec:training-freq}

The training frequency required depends crucially on the training {\em horizon} ($T_{horizon}$) for each model update. Prior work, Just-In-Time~\cite{mullapudi2019online}, trains the student model whenever it detects the accuracy has dropped below a threshold, and it trains only on the most recent frame (until the accuracy exceeds the threshold). This approach tends to overfit on recent frames, and therefore requires frequent retraining to maintain the desired accuracy.  While this is possible when training and inference occur on the same machine, it is impractical for \sys (\S\ref{sec:eval}).

Although lightweight models (e.g., those customized for mobile devices) have less capacity than large models, they can still generalize to some extent (e.g., over video frames captured in the same street, a specific room in a home, etc.). Therefore, rather than overfitting narrowly to one or a few frames, \sys uses a training horizon of several minutes. This reduces the required model update frequency, and helps mitigate sharp drops in accuracy when the model lags behind during scene changes (see~\cref{fig:cdf_diff}).

For semantic segmentation using DeeplabV3 with MobileNetV2~\cite{MobileNet-v2} backbone as the student model, we find that $T_{horizon} = 4$~minutes and $T_{update} = 10$~seconds work well across a wide variety of videos (\S\ref{sec:eval}). However, the optimal values of these parameters can depend on both the model capacity and the  video. For example, a lower-capacity student model might benefit from a shorter $T_{horizon}$ and $T_{update}$, and a stationary video with little scene change could use a longer $T_{update}$. Appendix~\ref{app:horizons_interplay} discusses the interplay between model capacity, $T_{horizon}$, and $T_{update}$ in more detail, and Appendix~\ref{app:atr} describes a simple technique to dynamically adapt $T_{update}$ to minimize bandwidth consumption (especially for stationary videos).

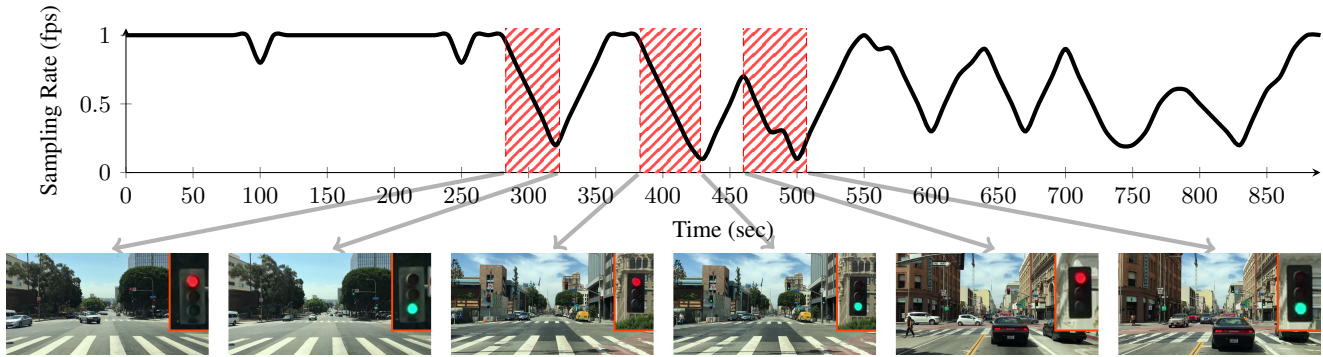
\begin{figure*}[!t]
\begin{tikzpicture}[smooth]
\begin{axis}[height=3.5cm, 
             width=1\linewidth,
             axis lines=left,
             ylabel near ticks,
             xlabel near ticks,
             xlabel={Time (sec)},
             ylabel={Sampling Rate (fps)},
             ymin=0,
             at={(0, 0)},
             ymax=1.05,
             font=\small,
             anchor=north east] (cpr)
        \fill[pattern color=red!70, pattern=flexible hatch,
        hatch distance=5pt,
        hatch thickness=1pt] (axis cs:283,0) rectangle (axis cs:323,1.1);
        \fill[pattern color=red!70, pattern=flexible hatch,
        hatch distance=5pt,
        hatch thickness=1pt] (axis cs:383,0) rectangle (axis cs:428,1.1);
        \fill[pattern color=red!70, pattern=flexible hatch,
        hatch distance=5pt,
        hatch thickness=1pt] (axis cs:460,0) rectangle (axis cs:507,1.1);
             \addplot [ultra thick, black, mark=none] table {figures/raw_data/adaptive_fps_timeseries.csv}; 
             \draw [dashed, color=red] (283,0) -- (283,110); 
             \draw [dashed, color=red] (323,0) -- (323,110); 
             \draw [dashed, color=red] (383,0) -- (383,110); 
             \draw [dashed, color=red] (428,0) -- (428,110); 
             \draw [dashed, color=red] (460,0) -- (460,110); 
             \draw [dashed, color=red] (507,0) -- (507,110);
        \node[inner sep=0pt] (p1) at (axis cs:283, 0){};
        \node[inner sep=0pt] (p2) at (axis cs:323, 0){};
        \node[inner sep=0pt] (p3) at (axis cs:383, 0){};
        \node[inner sep=0pt] (p4) at (axis cs:428, 0){};
        \node[inner sep=0pt] (p5) at (axis cs:460, 0){};
        \node[inner sep=0pt] (p6) at (axis cs:507, 0){};
\end{axis}
\node[inner sep=0pt, anchor=north east, outer sep=0] (lights) at (0,-3)
    {\includegraphics[width=\linewidth]{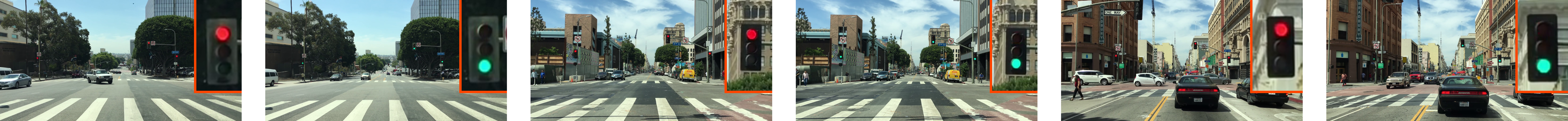}};
\begin{scope}[on background layer, shift={(lights.south west)},
                x={(lights.south east)},y={(lights.north west)}]
  \draw[->, color=black!30, ultra thick] (p1) -- (0.08, 1.05);
  \draw[->, color=black!30, ultra thick] (p2) -- (0.248, 1.05);
  \draw[->, color=black!30, ultra thick] (p3) -- (0.416, 1.05);
  \draw[->, color=black!30, ultra thick] (p4) -- (0.584, 1.05);
  \draw[->, color=black!30, ultra thick] (p5) -- (0.752, 1.05);
  \draw[->, color=black!30, ultra thick] (p6) -- (0.920, 1.05);
\end{scope}
\end{tikzpicture}
\caption{Adaptive frame sampling for a driving video. The sampling rate  decreases every time the car slows down for the red traffic light and increases as soon as the light turns green.}
\label{fig:asr}
\end{figure*}

\subsubsection{Which Parameters to Update?}
\label{sec:gradient-guided}

Na\"ively sending the entire student model to the edge device can consume significant bandwidth. For example, sending  DeeplabV3 with MobileNetV2 backbone, which  has 2 million (float16) parameters, every 10 seconds would require 3.2~Mbps of downlink bandwidth. To reduce bandwidth, we employ \emph{coordinate descent}~\cite{coordesc,Nesterov}, in which we train a small subset (e.g., 5\%) of parameters, $\mathcal{I}_n$, in each training phase $n$ and send only those parameters to the edge device.

To select $\mathcal{I}_n$, we use the  model gradients to identify the parameters (coordinates) that provide the largest improvement in the loss function when updated. A standard way to do this, called the \emph{\mbox{Gauss-Southwell} selection rule}~\cite{GG}, is to update the parameters with the largest gradient magnitude. We could compute the gradient for the entire model but only update the coordinates with the largest gradient values, leaving the rest of the parameters unmodified. This method works well for simple stateless optimizers like stochastic gradient descent (SGD), but optimizers like Adam~\cite{Adam} that maintain some internal state across training iterations require a more nuanced  approach. 

Adam keeps track  of moving averages of the first  and second  moments  of the gradient across training iterations. It uses this state to adjust the learning rate for each parameter dynamically based on the magnitude of ``noise'' observed in the gradients~\cite{Adam}.  Adam's internal state updates in each iteration depend on the point in the parameter space visited in that iteration. Therefore, to ensure the internal state is correct, we cannot simply compute Adam's updates for $K$ iterations, and then choose to keep only the coordinates with the largest change at the end. We must know {\em beforehand}  which coordinates we intend to update, so that we can update Adam's internal state consistently with the actual sequence of points visited throughout training.
\begin{algorithm}[t]
\caption{\emph{Gradient-Guided Method for Adam Optimizer}} \label{alg:coord_desc}
\footnotesize
\begin{algorithmic}[1]
    \STATE $\mathcal{I}_n \leftarrow$ Indices of $\gamma$ fraction with largest absolute values in $\bold{u}_{n-1}$ 
    \COMMENT{\textit{Entering $n^{th}$ Training Phase}}
    \STATE $\bold{b}_n \leftarrow$ binary mask of model parameters; 1 iff indexed by $\mathcal{I}_n$
    \STATE $\bold{w}_{n,0} \leftarrow \bold{w}_{n-1}$
    \COMMENT{Use the latest model parameters as the next starting point}
    \STATE $\bold{m}_{n,0} \leftarrow \bold{m}_{n-1,K}$
    \COMMENT{Initialize the first moment estimate to its latest value}
    \STATE $\bold{v}_{n,0} \leftarrow \bold{v}_{n-1,K}$
    \COMMENT{Initialize the second moment estimate to its latest value}
    \FOR {k $\in \{1,2,...,K\}$} 
    \STATE $\bold{S}_k \leftarrow$ Uniformly sample a mini-batch of data points from $\mathcal{B}$ over the last $T_{horizon}$ seconds 
    \STATE $\bold{g}_{n,k} \leftarrow \nabla_{\bold{w}} \tilde{\mathcal{L}}(\bold{S}_k; \bold{w}_{n,k-1})$ \COMMENT{Get the gradient of all model parameters w.r.t. loss on $\mathcal{S}_k$}
    \STATE $\bold{m}_{n,k} \leftarrow \beta_1 \cdot \bold{m}_{n,k-1} + (1-\beta_1) \cdot \bold{g}_{n,k}$ \COMMENT{Update first moment estimate}
    \STATE $\bold{v}_{n,k} \leftarrow \beta_2 \cdot \bold{v}_{n,k-1} + (1-\beta_2) \cdot \bold{g}_{n,k}^2$ \COMMENT{Update second moment estimate}
    \STATE $i  \leftarrow i + 1$ \COMMENT{Increment Adam's global step count}
    \STATE $\bold{u}_{n,k}  \leftarrow \alpha \cdot \frac{\sqrt{1-\beta_2^i}}{1-\beta_1^i}\cdot \frac{\bold{m}_{n,k}}{\sqrt{{\bold{v}}_{n,k} + \epsilon}}$ \COMMENT{Calculate the Adam updates for all model parameters}
    \STATE $\bold{w}_{n,k} \leftarrow \bold{w}_{n,k-1} - \bold{u}_{n,k} * \bold{b}_n$ \COMMENT{Update the parameters indexed by $\mathcal{I}_n$ \small{($*$ is elem.-wise mul.)}}
    \ENDFOR
    \STATE $\bold{u}_n \leftarrow \bold{u}_{n,K}$
    \STATE $\bold{w}_n \leftarrow \bold{w}_{n,K}$
\end{algorithmic}
\end{algorithm}

Our approach to coordinate descent for the Adam optimizer computes the subset of parameters that will be updated at the  start of each training phase, based on the coordinates that changed the most in the {\em previous} training phase. This subset is then fixed for the $K$   iterations of  Adam  in that training phase. 

The pseudo code in Algorithm~\ref{alg:coord_desc} describes the procedure in the $n^{th}$ training phase. Each training phase includes $K$ iterations with randomly-sampled mini-batches of data points from the last $T_{horizon}$ seconds of video. In iteration $k$, we update the first and second moments of the optimizer ($\bold{m}_{n,k}$ and $\bold{v}_{n,k}$) using the typical Adam rules (Lines 7--10). We then calculate the Adam updates for all model parameters $\bold{u}_{n,k}$ (Lines 11--12). However, we only apply the updates for parameters determined by the binary mask $\bold{b}_n$ (Line 13). Here, $\bold{b}_n$ is a vector of the same size as the model parameters, with ones at indices that are in $\mathcal{I}_n$ and zeros otherwise. We select the $\mathcal{I}_n$ to index the $\gamma$ fraction of parameters with the largest absolute value in the vector $\bold{u}_{n-1}$ (Line 1). We update $\bold{u}_{n}$ at the end of each training phase to reflect the latest Adam update for all parameters (Line 15). In the first training phase, $\mathcal{I}_n$ is selected uniformly at random. 

At the end of each training phase, the server sends the updated parameters $\bold{w}_n$ and their indices $\mathcal{I}_n$. For the indices, it sends a bit-vector identifying the location of the parameters. As the bit-vector is sparse, it can be compressed and we use gzip~\cite{gzip} in our implementation to carry this out. All in all, using gradient-guided coordinate descent to send 5\% of the parameters in each model update reduces downlink bandwidth by 13.3$\times$ with negligible loss in performance compared to updating the complete model (\S\ref{sec:results}).

\subsection{Reducing Uplink Bandwidth}\label{sec:design_uplink}

\sys adjusts the frame sampling rate at edge devices dynamically based on the extent and speed of scene change in a video. This helps reduce uplink (edge-to-server) bandwidth and server load for stationary or slowly-changing videos. 

To obtain a robust signal for scene change, we define a  metric, {\em $\phi$-score}, that tracks the rate of change in the  \emph{labels} associated with video frames. Compared to raw pixels, labels typically take values in a much smaller space (e.g., a few object classes), and therefore provide a more robust signal for measuring change. The server computes the $\phi$-score using the teacher model's labels. Consider a sequence of frames $\{\bold{I}_k\}_{k=0}^n$, and denote the teacher’s output on these frames by $\{\mathcal{T}(\bold{I}_k)\}_{k=0}^n$. For every frame $\bold{I}_k$, we define $\phi_k$ using the same loss function that defines the task, but computed using $\mathcal{T}(\bold{I}_k)$ and $\mathcal{T}(\bold{I}_{k-1})$ respectively to be the prediction and ground-truth labels. In other words, we set $\phi_k$ to be the loss (error) of the teacher model's prediction on ${I}_k$ with respect to the label $\mathcal{T}(\bold{I}_{k-1})$.
Hence, the smaller the $\phi_k$ score, the more alike are the labels for $\bold{I}_k$ and $\bold{I}_{k-1}$, i.e., stationary scenes tend to achieve lower scores.

The server measures the average $\phi$-score over recent frames, and periodically (e.g., every $\delta t = 10$~sec) updates the sampling rate at the edge device to try to maintain the $\phi$-score near a target value $\phi_{target}$:
\begin{equation}\label{eq:sampling_control}
    r_{t+1} = \Big[ r_t + \eta_r \cdot \big(\bar{\phi}_t - \phi_{target}\big) \Big]_{r_{min}}^{r_{max}},
\end{equation}
where $\eta_r$ is a step size parameter, and the notation $[\cdot]_{r_{min}}^{r_{max}}$ means the sampling rate is limited to the range $[r_{min}, r_{max}]$. We use  $r_{min} = 0.1$~fps (frames-per-second)  and $r_{max}=1$~fps in our implementation.

Fig.~\ref{fig:asr} shows an  example of adaptive sampling rate for a driving  video.  Notice how the sampling rate decreases when the car stops behind a red traffic light, and then increases once the light turns  green  and the car starts moving.

\para{Compression.} The edge device does not send sampled frames immediately. Instead it buffers samples corresponding to one model update interval ($T_{update}$, which the server communicates to the edge), and it runs H.264~\cite{H.264} video encoding on this buffer to compress it before transmission. The time taken at the edge device to fill the compression buffer and transmit a new batch of samples is hidden from the server by overlapping it with the training phase of the previous step. 
Performance isn't overly sensitive to the latency of delivering training data. As a result, it is possible to operate H.264 in a slow mode, achieving significant compression. Compressing the buffered samples in our experiments took at most 1 second.

\section{Evaluation}\label{sec:eval}
\subsection{Methodology} \label{subsec:method}
\para{Datasets.} We evaluate \sys on the task of semantic segmentation using four video datasets: Cityscapes~\cite{Cityscapes} driving sequence in Frankfurt (1 video, 46 mins long)\footnote{This video sequence is not labeled and was the only long video sequence available from Cityscapes (upon request).}, LVS~\cite{mullapudi2019online} (28 videos, 8 hours in total), A2D2~\cite{audivids} (3 videos, 36 mins in total), and  Outdoor Scenes (7 videos, 1.5 hours in total), which we collected from Youtube to cover a range of scene variability, including fixed cameras and moving cameras at walking, running, and driving speeds (see Appendix~\ref{app:datasets} for details and samples from Outdoor Scenes videos). 

\para{Metric.}\label{para:metric} To evaluate the accuracy of different schemes, we compare the inferred labels on the edge device with labels extracted for the same video frames using the teacher model.
For Cityscapes, A2D2, and Outdoor Scenes datasets we use DeeplabV3~\cite{Deeplab} model with Xception65~\cite{Xception65} backbone (2048$\times$1024 input resolution) trained on the Cityscapes dataset~\cite{Cityscapes} as the teacher model. For LVS, we follow Mullapudi~\etal~\cite{mullapudi2019online} in using Mask~R-CNN~\cite{he2017mask} trained on the MS-COCO dataset~\cite{MS-COCO} as the teacher model. Labeling each frame using the teacher models takes 200--300ms on a V100 GPU. We report the mean Intersection-over-Union (mIoU) metric relative to the labels produced by this reference model. The metric computes the Intersection-over-Union (defined as the number of true positives divided by the sum of true positives, false negatives and false positives) for each class, and takes a mean over the classes. We manually select a subset of most common output classes in each of these videos as summarized in~\cref{table:dataset} in~\cref{app:datasets}.

\begin{table*}[ht!]
\centering
\setlength\tabcolsep{4pt}
\small
\begin{tabular}{l l c c c c c }
\hline
\textbf{Dataset} & \textbf{Metric} & \textbf{\frozen} & \textbf{\early} & \textbf{\remoteinference} & \textbf{\jitcustom} & \textbf{\sys} \\
\hline
\hline
\multirow{2}{*}{\textbf{Outdoor Scenes~}} & mIoU (\%) & 63.68 & 69.73 & 69.05 & 73.14 & \textbf{74.26}\\
& Up/Down BW (Kbps) & 0/0 & 63.1/91.4 & 1949/54.6 & 2735/3109 & 189/205\\
\hline
\multirow{2}{*}{\textbf{A2D2~\cite{audivids}}} & mIoU (\%) & 62.05 & 50.78 & 63.25 & 69.23 & \textbf{69.31}\\
& Up/Down BW (Kbps) & 0/0 & 56.9/100 & 1927/40.5 & 2487/2872 & 158/203\\
\hline
\multirow{2}{*}{\textbf{Cityscapes~\cite{Cityscapes}}} & mIoU (\%) & 73.08 & 63.90 & 66.53 & \textbf{75.75} & 75.66\\
& Up/Down BW (Kbps) & 0/0 & 8.2/49.2 & 1667/50.8 & 2920/3294 & 164/226\\
\hline
\multirow{2}{*}{\textbf{LVS~\cite{mullapudi2019online}}} & mIoU (\%) & 59.32 & 64.88 & 61.52 & 65.70 & \textbf{67.38}\\
& Up/Down BW (Kbps) & 0/0 & 48.1/77.4 & 1865/21.6 & 2456/3264 & 165/207\\
\hline
\end{tabular}
\caption{Comparison of \gls{miou} (in percent), Uplink and Downlink bandwidth (in Kbps) for different methods across 4 video datasets.}
\label{table:benchmark_dataset}
\end{table*}

\para{Schemes.} On the edge device, we use the DeeplabV3 with MobileNetV2~\cite{MobileNet-v2} backbone at a 512$\times$256 input resolution, which runs smoothly in real-time at 30 \gls{fps} on a Samsung Galaxy S10+ phone's Adreno~640~GPU with less than 40 ms camera-to-label latency.

 We compare the following schemes:
\begin{itemize}[noitemsep, topsep=0pt, leftmargin=10pt]
    \item \textbf{\frozen:} We run a pre-trained model on the edge device without video-specific customization. For the LVS dataset, we use a checkpoint pre-trained on PASCAL VOC 2012 dataset~\cite{miou}. For the rest of the datasets we used a checkpoint pre-trained for Cityscapes~\cite{Cityscapes}.
    \item \textbf{\early:} We fine-tune the entire model on the first 60 seconds of the video at the server and send it to the edge. This adaptation happens only once for every video. Comparing this scheme with \sys will show the benefit of \emph{continuous} adaptation. 
    \item \textbf{\remoteinference:} We use the teacher model at a remote server to infer the labels on sample frames (one frame every second), which are then sent to the device. The device locally interpolates the labels to 30 frames-per-second using optical flow tracking~\cite{zhu2017deep,beauchemin1995computation}. For tracking, we use the OpenCV implementation of Farneback optical flow  estimation~\cite{farneback2003two} with 5 iterations, Gaussian filters of size 64, 3 pyramid levels, and a polynomial degree of 5 at 1024$\times$512 resolution. Although it takes 700 ms to compute the flows for each frame in our tests on a Linux CPU machine, we assumed an optimized implementation with edge hardware support can run in real-time~\cite{nvidia_opticalflow} in favor of this approach. We set the sampling rate to 1~fps, which matches the the maximum sampling rate for \sys. Note that, unlike \sys, this approach cannot apply the ``buffer compression'' method (see~\S\ref{sec:design_uplink}) as the buffering latency would make the labels stale. To avoid accuracy loss, we send the samples at full quality with this scheme; this requires about 2 Mbps of uplink bandwidth.\footnote{For reference, sending one frame per second with a good JPEG quality (quality index of 75) at this resolution requires $\sim$700 Kbps of bandwidth.}
    \item \textbf{\jitcustom:} We deploy the online distillation algorithm proposed by~\cite{mullapudi2019online} at the sever. This scheme trains the student model on the most recent sample frame until its training accuracy meets a threshold. Using the default parameters, it increases the sampling/training frequency (up to one model update every 266 ms) if it cannot meet the threshold accuracy within a maximum number of training iterations. Mullapudi~\etal~\cite{mullapudi2019online} also propose a specific lightweight model, JITNet. However, their Just-In-Time adaptation algorithm is general and can be used with any model. We evaluated Just-In-Time training with both our default student model (DeeplabV3 with MobileNetV2 backbone) and the JITNet architecture, and found they achieve similar performance in terms of both accuracy (less than 2 percent difference in mIoU) and number of model updates.\footnote{Our implementation of the JITNet model on Samsung Galaxy S10+ mobile CPU runs 2$\times$ slower at the same input resolution compared to DeeplabV3 with MobileNetV2 backbone.} Hence, we report the results of this approach for the same model as AMS for a more straightforward comparison. Similar to \sys, we use the gradient-guided strategy (\S\ref{sec:gradient-guided}) for this scheme to adapt 5\% of the model parameters in each update, which actually achieves a slightly better overall performance (e.g., 1.2\% mIoU increase on Outdoor Scenes dataset) than updating the entire model. 
    The accuracy threshold is a controllable hyper-parameter that determines the frequency of model updates. A higher threshold achieves better accuracy at the cost of higher downlink bandwidth for sending model updates. We set the accuracy threshold to achieve roughly the same accuracy as \sys on each video, allowing us to compare their bandwidth usage at the same accuracy. Using Just-In-Time's default threshold (75\%) improves overall accuracy by 1.0\% at the cost of 3.3$\times$ higher bandwidth. Following~\cite{mullapudi2019online}, we use the Momentum Optimizer~\cite{QIAN1999145} with a momentum of $0.9$. 
    \item \textbf{\sys:} We use Algorithm~\ref{alg:server_training} at the server with $T_{horizon}=240$~sec, and $K=20$ iterations. 
    We set the ASR parameters $r_{min}$ and $r_{max}$ to 0.1 and 1 frames-per-seconds respectively, with $\delta t=10$~sec.
    Unless otherwise stated, 5\% of the model parameters are selected using the gradient-guided strategy. In the uplink, we compress and send the buffer of sampled frames described in \S\ref{sec:design_uplink} using H.264 in the two-pass mode at medium preset speed and a target bitrate of 200 Kbps. 
    We used AMS with the same set of hyper-parameters for all 39 videos across the four datasets. For training, we use Adam optimizer~\cite{Adam} with a learning rate of $0.001$ ($\beta_1=0.9$, $\beta_2=0.999$).
\end{itemize}

We use a single NVIDIA Tesla V100 GPU at the server for all schemes. To better understand the behavior of different schemes in terms of system requirements, we measure the resource usages under no significant network limitations. We defer a study of each scheme's behaviour under resource contention to future work.

\begin{table}[t!]
\small
\centering
\setlength\tabcolsep{2pt}
\begin{tabular}{ l c c c c c }
\hline
\textbf{Description} & \textbf{No Cust.} & \textbf{\early} & \textbf{Rem.+Trac.} & \textbf{JIT} & \textbf{\sys} \\
\hline 
\hline
Interview  & 71.91 & 87.40 & \textbf{89.98} & 86.47 & 87.75\\
Dance recording & 72.80 & 84.26 & \textbf{86.41} & 84.40 & 83.88 \\
Street comedian  & 54.49 & 65.06 & 58.81 & 69.79 & \textbf{72.03}\\
Walking in Paris  & 69.94 & 67.63 & 69.59 & 75.22 & \textbf{75.87 }\\
Walking in NYC  & 49.05 & 54.96 & 54.49 & 56.54 & \textbf{59.74} \\ 
Driving in LA  & 66.26 & 66.30 & 66.48 & 70.95 & \textbf{71.01} \\
Running & 61.32 & 62.51 & 57.57 & 68.64 & \textbf{69.55} \\
\hline
\end{tabular}
\caption{Impact of the scene variations pace on \gls{miou} (in percent) for different methods across the videos in Outdoor Scenes dataset. 
}
\label{table:benchmark_outdoor}
\end{table}

\subsection{Results}\label{sec:results}

\para{Comparison to baselines.} \Cref{table:benchmark_dataset} summarizes the results across the four datasets. We report the mIoU, uplink and downlink bandwidth, averaged over the videos in each dataset. We also report per-video results for the Outdoor Scenes dataset in \Cref{table:benchmark_outdoor}. The main takeaways are: 
\begin{enumerate}[noitemsep, topsep=0pt, leftmargin=10pt]
    \item Adapting the edge model provides significant mIoU gains.
    \sys achieves 0.4--17.8\% (8.3\% on average) better mIoU score than \frozen.
    
    \item \early is sometimes better and sometimes worse than \frozen. Recall that \early specializes the model based on the first minute of a video. When the first minute is representative of the entire video, \early can improve accuracy. However, on videos that vary significantly over time (e.g., driving scenes in A2D2 and Cityscapes), customizing the model for the first minute can backfire. By contrast, \sys consistently improves accuracy (up to 39.1\% for some videos, 4.3\% on average compared to \early) since it continually adapts the model to video dynamics.

    \item \remoteinference performs better on static videos since optical flow tracking works better in these cases. However, it struggles on more dynamic videos and performs worse than \sys (up to 24.4\% on certain videos, 5.8\% on average). For example, note that in~\cref{table:benchmark_outdoor}, \remoteinference performs no better than \frozen (which does not use the network) on the Driving in LA, Walking in Paris, and Running videos. \remoteinference requires much less bandwidth in the downlink compared to Just-In-Time and \sys as it downloads labels rather than model updates. However, in the uplink it requires about 2Mbps of bandwidth since it cannot buffer and compress frames to ensure it receives labels with low latency (unlike \sys).
    
    \item \jitcustom achieves the closest overall mIoU score to \sys, but it requires 4.4--44.5$\times$ more downlink bandwidth (15.7$\times$ on average), and 5.2--37.1$\times$ more uplink bandwidth (16.8$\times$ on average) across all videos. Across all videos, \sys requires only 181--225 Kbps downlink bandwidth and 57--296 Kbps uplink bandwidth.

\end{enumerate}

\begin{figure}[!t]
    \centering
    {
    {\trimbox{5pt 3pt 5pt 5pt}{%
    \begin{tikzpicture}[mark size=3pt]%
    \begin{semilogxaxis}[
    name=MyAxis,
    height=5cm,
    axis lines=left,
    axis line style = thick,
    enlargelimits=true,
    width=\linewidth,
    log basis x={10},
    xlabel=\textbf{Average Downlink BW (Kbps)},
    ylabel=\textbf{mIoU (\%)},
    font=\small,
    legend columns=1,
                     legend style={at={(0.63,0.17)}, 
                           anchor=west ,fill=white, fill opacity=0.5, draw opacity=1,text opacity=1, draw=none, font=\small, inner sep=0},
                           legend cell align={left},
                           grid=both,
                          ymax=80,
                          ymin=58,
                          xmin=50,
                           ylabel near ticks,
    ]
    \addlegendimage{mark=o, only marks};
    \addlegendimage{mark=square, only marks};
    \input{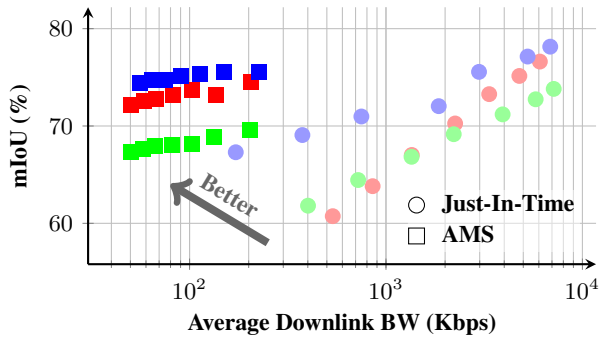}
    \legend{\ \ \textbf{\jitcustom}\ , \ \ \textbf{\sys} };
    \draw[->, color=black!60, line width=3pt] (axis cs:250,58) -- (axis cs:80,64) node [midway, above, sloped,font=\small, color=black!60, fill=none] (TextNode) {\textbf{Better}};
    \end{semilogxaxis}
    \end{tikzpicture}}}}
    \caption{mIoU vs. downlink bandwidth for \sys and \jitcustom with different parameters. Each color represents one dataset and each marker's shape/tint represents the scheme.}
        \vspace{-14pt}
    \label{fig:acc_vs_bw}
\end{figure}

\para{Impact of AMS and Just-In-Time parameters.} Both \sys and \jitcustom have parameters that affect their accuracy and model-update frequency. To compare these schemes more comprehensively, we sweep these parameters and measure the mIoU and downlink bandwidth they achieve at each operating point.  For \jitcustom, we vary the target accuracy threshold in the interval 55--85 percent, and for \sys, we vary $T_{update}$ between 10 to 40 seconds. \cref{fig:acc_vs_bw} shows the results for 3  datasets (Cityscapes, A2D2, and Outdoor Scenes).\footnote{We omit the LVS dataset from these results to reduce cost of running the experiments in the cloud.} Comparing the data points of the same color (same dataset) for the two schemes, we observe that \jitcustom requires about 10$\times$ more bandwidth to achieve the same accuracy as \sys. Note that we apply our gradient-guided parameter selection to \jitcustom; without this, it would have required 150$\times$ more bandwidth than \sys. \sys is less sensitive to limited bandwidth than \jitcustom (notice the difference in slope of mIoU vs. bandwidth for the two schemes). As discussed in \S\ref{sec:training-freq}, the reason is that \sys trains the student model over a longer time horizon (as opposed to a single recent frame). Thus it generalizes better and can tolerate fewer model updates more gracefully.

\para{Impact of the gradient-guided method.} \Cref{tab:coordinate-descent} compares the \gstrat method descibed in \S\ref{sec:gradient-guided} with other approaches for selecting a subset of parameters (coordinates) in the training phase on the Outdoor Scene dataset. The \emph{First}, \emph{Last}, and \emph{First\&Last} methods select the parameters from the initial layers, final layers, and split equally from both, respectively. \emph{Random} samples parameters uniformly from the entire network.
Gradient-guided performs best, followed by Random. Random is notably worse than \gstrat when training a very small fraction (1\%) of model parameters. The methods that update only the first or last model layers are substantially worse than the other approaches. 

Overall, Table~\ref{tab:coordinate-descent} shows that \sys's gradient-guided method is very effective. Sending only 5\% of the model parameters results in only 0.73\% loss of accuracy on average (on the Outdoor Scenes dataset), but it reduces the downlink bandwidth requirement from 3.3~Mbps for full-model updates to 205~Kbps.

\begin{table}
    \centering
\small
\setlength\tabcolsep{8pt}
\begin{tabularx}{\linewidth}{lcccc}
\toprule
& \multicolumn{4}{c}{\textbf{Fraction}}\\
\cmidrule{2-5}
\textbf{Strategy} & 20\% & 10\% & 5\% & 1\% \\
\midrule
Last Layers & -5.98 & -6.58 & \cellcolor{lightgray}{-8.98} & -10.99\\
First Layers & -2.63 & -5.54 & \cellcolor{lightgray}{-8.37} & -15.45\\
First\&Last Layers & -1.0 & -2.29 & \cellcolor{lightgray}{-3.54} & -7.30\\ 
Random Selection& -0.21 & -0.70 & \cellcolor{lightgray}{-2.90} & -5.29\\
Gradient-Guided & +0.13 & -0.13 & \cellcolor{lightgray}{-0.73} & -2.87\\ 
\midrule
BW (Kbps) & 715 & 384 & \cellcolor{lightgray}{205} & 46\\
\multicolumn{2}{l}{\textbf{Full model BW (Kbps)}} & \textbf{3302} & &\\
\bottomrule
\end{tabularx}
\caption{Average difference in mIoU relative to full-model training (in percent) for different coordinate descent strategies on the Outdoor Scenes dataset.}
\label{tab:coordinate-descent}
\end{table}

\begin{figure}[!t]
    \centering
    \trimbox{0.3cm 0.1cm 0.3cm 0.2cm}{%
    \begin{tikzpicture}[smooth]
        \begin{axis}[height=5cm,
                     width=\linewidth,
                     axis lines=left,
                     xlabel={\textbf{\gls{miou} Gain over \frozen Scheme (\%)}},
                     ylabel={\textbf{CDF}},
                     font=\small,
                     legend columns=1,
                     enlargelimits=true,
                     legend style={at={(0.03,0.7)}, 
                          anchor=west, fill=white, fill opacity=0.7, draw opacity=1, 
                          text opacity=1, draw=none, font=\small, inner sep=0},
                          legend cell align={left},
                     xmax=40,
                     xmin=-40,
                     ymax=1, 
                     ymin=-0,
                     ytick={0, 0.2, 0.4, 0.6, 0.8, 1},
                     ylabel near ticks,
                     xlabel near ticks,
                     grid=both]
            \input{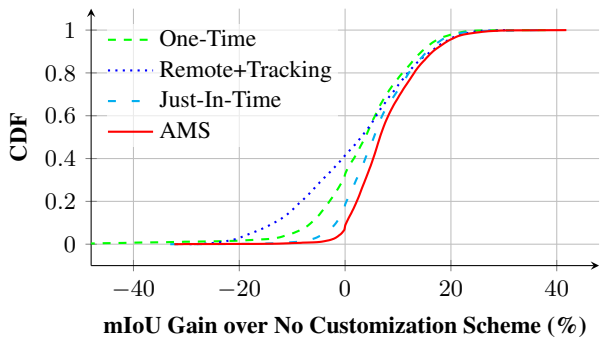}
            \legend{\early, \remoteinference, \jitcustom, \sys};
        \end{axis}
    \end{tikzpicture}}
    \caption{CDF of \gls{miou} gain relative to \frozen across all frames for different schemes.}
            \vspace{-8pt}
    \label{fig:cdf_diff}
\end{figure}

\para{Robustness to scene changes.} 
Does \sys consistently improve accuracy across all frames or are the benefits limited to certain segments of video with stationary scenes? To answer this question, \cref{fig:cdf_diff} plots the cumulative distribution of mIoU improvement relative to \frozen across all frames (more than 1 million frames across the four datasets) for all schemes. \sys consistently outperforms the other schemes. Surprisingly, Just-In-Time has worse accuracy than \sys, despite updating its model much more frequently. \sys achieves better mIoU than \frozen in 93\% of frames, while Just-In-Time and One-Time customization are only better than \frozen 82\% and 67\% of the time. This shows that \sys's training strategy, which avoids  overfitting to a few recent frames, is more robust and handles scene variations better.

\begin{figure}
\begin{tikzpicture}
                \begin{axis}[height=4.5cm, 
                         width=1\linewidth,
                         axis lines=left,
                         ylabel near ticks,
                         xlabel near ticks,
                         xlabel=\textbf{Number of clients},
                         ylabel=\textbf{$\Delta$ \gls{miou} (\%)},
                         grid=both,
                         legend style={at={(0,0)}, 
                           anchor=south west ,fill=white, fill opacity=0, draw opacity=1,text opacity=1, draw=none, font=\small},
                         xmax=10,
                         ymin=-2,]
                   \addplot+[ultra thick, violet] coordinates{
                    (1, 0.000000)
                    (2, 0.000000)
                    (3, 0.000000)
                    (4, -0.054144)
                    (5, -0.176154)
                    (6, -0.340016)
                    (7, -0.531612)
                    (8, -0.771514)
                    (9, -1.058779)
                    (10, -1.383912)};
                \end{axis}
            \end{tikzpicture}
            \caption{Average multiclient \gls{miou} degradation compared to single-client performance on Outdoor Scenes dataset.}%
            \label{fig:multi_client}
            \end{figure}
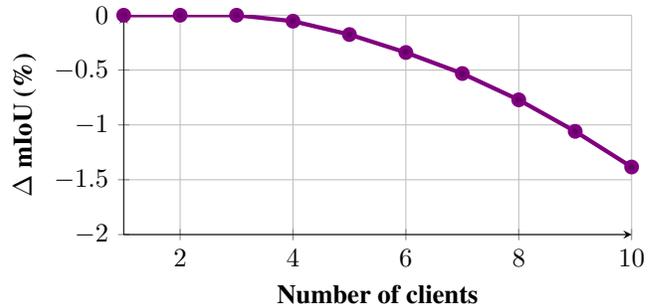
            
\para{Multiple edge devices.}\label{para:multiclient} 
\Cref{fig:multi_client} show the accuracy degradation (w.r.t. single edge device) when multiple edge device share a single GPU at the server in round-robin manner. By giving more GPU time to videos with more scene variation, \sys scales to supporting up to 9 edge device on a single V100 GPU at the server with less than 1\% loss in \gls{miou} (see Appendix~\ref{app:atr} for more details). 

\section{Conclusion}
We presented \sys, an approach for improving the performance of real-time video inference on low-powered edge devices that uses a remote server to continually train and stream model updates to the edge device. Our design centers on reducing communication overhead: avoiding excessive overfitting, updating a small fraction of model parameters, and adaptively sampling training frames at edge devices. \sys makes over-the-network model adaptation possible with a few 100~Kbps of uplink and downlink bandwidth, levels easily sustainable on today's (wireless) networks. Our results showed that \sys improves accuracy of semantic segmentation using a mobile-friendly model by 0.4--17.8\% compared to a pretrained (uncustomized) model across a variety of videos, and requires 15.4$\times$ less bandwidth to achieve similar accuracy to recent online  distillation methods.

{\small
\bibliographystyle{ieee_fullname}
\bibliography{reference}
}

\clearpage
\appendix
\begin{appendices}
\section{Videos}\label{app:datasets}
 \begin{table*}
\small
\centering
\begin{tabular}{ c l l }
\toprule
\textbf{Dataset} & \textbf{Description} & \textbf{Classes} \\
\midrule
\multirow{7}{*}{\textbf{Outdoor Scenes}} & Interview & Building, Vegetation, Terrain, Sky, Person, Car \\
\cmidrule{2-3}
 & Dance Recording & Sidewalk, Building, Vegetation, Sky, Person \\
\cmidrule{2-3}
 & Street Comedian & Road, Sidewalk, Building, Vegetation, Sky, Person \\
\cmidrule{2-3}
 & Walking in Paris & Road, Building, Vegetation, Sky, Person, Car \\
\cmidrule{2-3}
 & Walking in NY & Road, Building, Vegetation, Sky, Person, Car \\
\cmidrule{2-3}
 & Driving in LA & Road, Sidewalk, Building, Vegetation, Sky, Person, Car \\
\cmidrule{2-3}
 & Running & Road, Vegetation, Terrain, Sky, Person \\
\midrule
\multirow{3}{*}{\textbf{A2D2}~\cite{audivids}} & Driving in Gaimersheim & Road, Sidewalk, Building, Sky, Person, Car \\
\cmidrule{2-3}
 & Driving in Munich & Road, Sidewalk, Building, Sky, Person, Car \\
\cmidrule{2-3}
 & Driving in Ingolstadt & Road, Sidewalk, Building, Sky, Person, Car \\
\midrule
\textbf{Cityscapes~\cite{Cityscapes}} & Driving in Frankfurt & Road, Sidewalk, Building, Sky, Person, Car \\
\midrule
\multirow{29}{*}{\textbf{LVS}~\cite{mullapudi2019online}} & Badminton & Person \\
\cmidrule{2-3}
 & Squash & Person \\
\cmidrule{2-3}
 & Table Tennis & Person \\
\cmidrule{2-3}
 & Softball & Person \\
\cmidrule{2-3}
 & Hockey & Person \\
\cmidrule{2-3}
 & Soccer & Person \\
\cmidrule{2-3}
 & Tennis & Person \\
\cmidrule{2-3}
 & Volleyball & Person \\
\cmidrule{2-3}
 & Ice Hockey & Person \\
\cmidrule{2-3}
 & Kabaddi & Person \\
\cmidrule{2-3}
 & Figure Skating & Person \\
\cmidrule{2-3}
 & Drone & Person \\
\cmidrule{2-3}
 & Birds & Bird \\
\cmidrule{2-3}
 & Dogs & Car, Dog, Person \\
\cmidrule{2-3}
 & Horses & Horse, Person \\
\cmidrule{2-3}
 & Ego Ice Hockey & Person \\
\cmidrule{2-3}
 & Ego Basketball & Car, Person \\
\cmidrule{2-3}
 & Ego Dodgeball & Person \\
\cmidrule{2-3}
 & Ego Soccer & Person \\
\cmidrule{2-3}
 & Biking & Bicycle, Person \\
\cmidrule{2-3}
 & Streetcam1 & Car, Person \\
\cmidrule{2-3}
 & Streetcam2 & Car, Person \\
\cmidrule{2-3}
 & Jackson Hole & Car, Person \\
\cmidrule{2-3}
 & Murphys & Bicycle, Car, Person \\
\cmidrule{2-3}
 & Samui Street & Bicycle, Car, Person \\
\cmidrule{2-3}
 & Toomer & Car, Person \\
\cmidrule{2-3}
 & Driving & Bicycle, Car, Person \\
\cmidrule{2-3}
 & Walking & Bicycle, Car, Person \\
\bottomrule
\end{tabular}
\caption{Summary of the video datasets and their target classes in evaluations.}
\label{table:dataset}
\end{table*}

\subsection{Outdoor Scenes Dataset}
\begin{figure*}
    \centering
    \includegraphics[width=\textwidth]{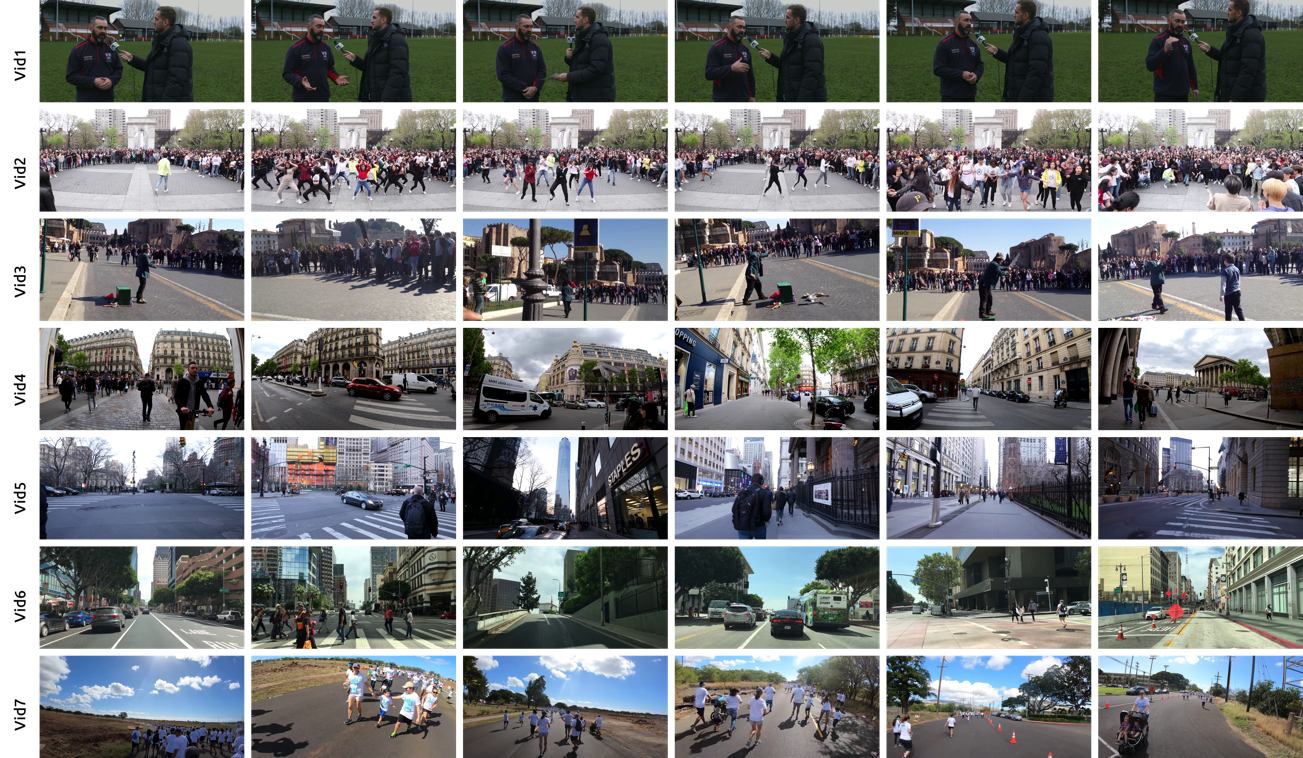}
    \caption{Sample video frames. Rows (from top to bottom) correspond to Interview, Dance Recording, Street Comedian, Walking in Paris, Walking in NY, Driving in LA, and Running.}
    \label{fig:bare_frame_samples}
 \end{figure*}

The Outdoor Scenes video dataset includes seven publicly available videos from Youtube, with 7--15 minutes in duration. These videos span different levels of scene variability and were captured with four types of cameras: Stationary, Phone, Headcam, and DashCam.  For each video, we manually select 5--7 classes that are detected frequently by our best semantic segmentation model (DeeplabV3 with Xception65 backbone trained on Cityscapes data) at full resolution. \Cref{table:dataset} shows summary information about these videos. In~\cref{fig:bare_frame_samples} we show six sample frames for each video. For viewing the raw and labeled videos, please refer to \href{https://github.com/modelstreaming/ams}{https://github.com/modelstreaming/ams}.

\subsection{Prior Work Videos}
In our experiments, we also evaluate \sys on three long video datasets from prior work: Cityscapes~\cite{Cityscapes} driving sequence in Frankfurt (1 video, 46 mins long)\footnote{This video sequence is not labeled and was the only long video sequence available from Cityscapes (upon request).}, LVS~\cite{mullapudi2019online} (28 videos, 8 hours in total), A2D2~\cite{audivids} (3 videos, 36 mins in total). \Cref{table:dataset} shows the summary information of the classes present in each video in these datasets. Overall, LVS includes fewer classes per video, and A2D2 and Cityscapes only include driving scenes. Hence, we introduced the Outdoor Scenes dataset that includes more diverse scenes and more classes.

\section{Other Related Work}\label{app:related}
\para{Continual/Lifelong Learning.} The goal of lifelong learning~\cite{lopez2017gradient} is to accumulate knowledge over time~\cite{rolnick2019experience}. Hence the main challenge is to improve the model based on new data over time, while not forgetting data observed in the past~\cite{kirkpatrick2017overcoming,mccloskey1989catastrophic}. However, as the lightweight models have limited capacity,  in \sys we aim to track the best model at each point in time, and these models are allowed not to have the same performance on the old data. 

\para{Meta Learning.} Meta learning~\cite{finn2017model,ravi2017optimization,finn2019online} algorithms aim to learn models that can be adapted to any target task in a set of tasks, given only few samples (shots) from that task. Meta learning is not a natural framework for continual model specialization for video. First, as videos have temporal coherence, there is little benefit in handling an arbitrary order of task arrival. Indeed, it is more natural to adapt the latest model over time instead of always training from a meta-learned initial model.\footnote{An exception is a video that changes substantially in a short period of time, for example, a camera that moves between indoor and outdoor environments. In such cases, a meta model may enable faster adaptation.} Second, training such a meta model usually requires two nested optimization steps~\cite{finn2017model}, which would significantly increase the server's computation overhead. Finally, most meta learning work considers a finite set of tasks but this notion is not well-suited to video. 

\para{Federated Learning.} Another body of work on improving edge models over time is federated learning~\cite{mcmahan2017communication}, in which the training mainly happens on the edge devices and device updates are aggregated at a server. The server then broadcasts the aggregated model back to all devices. Such updates happen at a time scale of hours to days~\cite{bonawitz2019towards}, and they aim to learn better generalizable models that incorporate data from all edge devices. In contrast, the training in \sys takes place at the server at a time scale of a couple of seconds and intends to improve the accuracy of an individual edge device's model for its particular video. 

\para{Unsupervised Adaptation.} Domain adaptation methods~\cite{ben2007analysis,kang2019contrastive} intend to compensate for shifts between training and test data distributions. In a typical approach, an unsupervised algorithm fine-tunes the model over the entire test data at once. However, frame distribution can drift over time. As our results show, one-time fine-tuning can have a much lower accuracy than continuous adaptation. However, it is too expensive to provide a fast adaptation (at a timescale of 10--100 seconds) of these models at the edge, even using unsupervised methods. Moreover, using \sys we benefit from the superior performance of supervised training by running state-of-the-art models as the ``teacher'' for knowledge distillation~\cite{hinton2015distilling} in the cloud. 

\section{Impact of model capacity and training horizon on online adaptation}\label{app:horizons_interplay}
AMS improves the performance of a lightweight model on edge devices through continual online adaptation. There exists a tradeoff in this approach between boosting the model's accuracy for the current data  distribution and overfitting, which degrades performance when the data distribution changes. This tradeoff depends on the nature of the video (how fast the scenes change) and the model capacity. For instance, a low-capacity model may perform better with frequent updates despite overfitting.  

Prior work on online  model adaptation for  video largely targets the overfitting regime: Just-In-Time~\cite{mullapudi2019online} trains the lightweight model when it detects accuracy has dropped below a threshold, and it focuses its training on boosting the model's performance on the most recent frames. Just-In-Time must therefore repeatedly retrain its model as scenes change to  maintain the desired accuracy.  While this approach is sensible when training and inference both  occur on the same powerful machine, it is impractical for online model training at a remote server. As our evaluation results  show (\S\ref{sec:results}), Just-In-Time's approach requires very frequent model updates, incurring a high communication overhead.

We seek to avoid the need for frequent model updates by training the model over a suitable time horizon\,---\,not too short (which can lead to perpetual overfitting), but also not too long (which can hurt accuracy). The key observation is  that although practical lightweight models (e.g., those  customized for mobile devices) lack the generalization capacity of state-of-the-art models, they still have adequate capacity to generalize over a narrower distribution of frames (e.g., video captured in the same street, a specific room in a house,  etc.).    

To illustrate these issues, consider the model adaptation framework described in \S\ref{sec:framework} with two knobs:  (i) $T_{update}$, the {\em model update interval};  each $T_{update}$ seconds the model is trained and updated. (ii) $T_{horizon}$, the {\em training horizon}; each update uses (sampled) frames from the last $T_{horizon}$ seconds of video to train the model. For effective model adaptation, these two knobs are inter-dependent. When  $T_{horizon}$ is   small,  the model updates tend to overfit  and therefore must be frequent (small $T_{update}$),  while  a larger $T_{horizon}$ can produce models that generalize better and maintain high accuracy  over a larger $T_{update}$ interval. Picking too  large of a $T_{horizon}$, however, is also not ideal, as the model might not have sufficient capacity to generalize over a wide  distribution of frames, reducing its accuracy. 

\begin{figure}
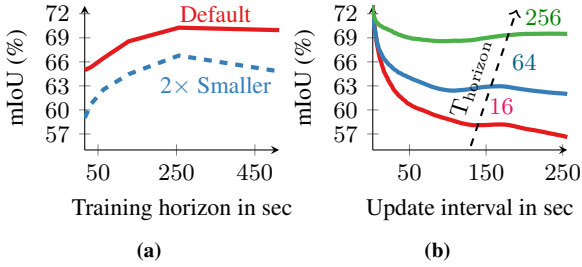

    \subfloat[\label{fig:train_horizon}]{%
    {\trimbox{0.2cm 0cm 0.2cm 0}{%
    \begin{tikzpicture}%
        \begin{axis}[height=3.5cm, 
                     width=0.5\linewidth,
                     axis lines=left,
                     ylabel near ticks, 
                     xlabel near ticks,
                     xlabel={Training horizon  in sec},
                     ylabel={\gls{miou} (\%)},
                     legend columns=-1,
                     legend style={at={(0.35,1.15)}, 
                           anchor=center ,fill=white, fill opacity=0, draw opacity=1,text opacity=1, draw=none, font=\small},
                     xmax=512,
                     ymax=73,
                     ymin=55,
                     xtick={50,250,450},
                     ytick={57, 60, 63, 66, 69, 72},
                     font=\small]
            \input{figures/raw_data/k1_full}
            \input{figures/raw_data/k1_0.5}
            \node[] at (axis cs:350,72) {\textcolor{Set1-A}{Default}};
            \node[] at (axis cs:350,63) {\textcolor{Set1-B}{2$\times$ Smaller}};
        \end{axis}
    \end{tikzpicture}}}}
    \subfloat[\label{fig:retrain_freq}]{%
    {\trimbox{0cm 0cm 0cm 0}{%
        \begin{tikzpicture}%
        \begin{axis}[height=3.5cm, 
                     width=0.5\linewidth,
                     axis lines=left,
                     ylabel near ticks, 
                     xlabel near ticks,
                     xlabel={Update interval in sec},
                     ylabel={\gls{miou} (\%)},
                     legend columns=-1,
                     legend style={at={(0.35,1.15)}, 
                           anchor=center ,fill=white, fill opacity=0, draw opacity=1,text opacity=1, draw=none, font=\small},
                     xmax=256,
                     ymin=55,
                     ymax=73,
                     xtick={50,150,250},
                     ytick={57, 60, 63, 66, 69, 72},
                     font=\small]
            \input{figures/raw_data/k1_16}
            \input{figures/raw_data/k1_64}
            \input{figures/raw_data/k1_256}
            \node[rotate=72] at (axis cs:130, 64){$\mathrm{T_{horizon}}$};
            \draw[->, dashed, color=black, thick] (axis cs:130, 55.5) -- (axis cs:190, 72.5);
            \node[] at (axis cs:170,60.5) {\textcolor{OrangeRed}{$16$}};
            \node[] at (axis cs:200,66) {\textcolor{MidnightBlue!75!black}{$64$}};
            \node[] at (axis cs:225,71.5) {\textcolor{green!50!black}{$256$}};
        \end{axis}
    \end{tikzpicture}}}}%
    \captionof{figure}{Impact of training horizon and model update  interval on the mean-intersection-over-union (mIoU) accuracy for semantic segmentation.}%
    \label{fig:adapt_intuition}
\end{figure}

Figure~\ref{fig:adapt_intuition} illustrates this intuition for the video semantic segmentation task. We consider two variations of the lightweight model: {\em (i)} DeeplabV3 with MobileNetV2 backbone, {\em (ii)} a smaller version with the same architecture but with half the number of channels in each convolutional layer. We pick 50 points in time uniformly distributed over the course of a video of driving scenes  in Los Angeles. For each time $t$, we train the two lightweight models on the frames in the interval $[t-T_{horizon}, t)$ and then evaluate them on frames in $[t,t+T_{update})$ (with $T_{update}=16$ sec). 

We plot the average accuracy of the two variants for each value of $T_{horizon}$ in Fig.~\ref{fig:train_horizon}. As expected, the smaller model's accuracy peaks at some training horizon ($T_{horizon} \approx 256$ sec) and degrades with larger $T_{horizon}$ as the model capacity becomes insufficient. The default model exhibits a similar behavior, but with a more gradual drop for large $T_{horizon}$. 
Fig.~\ref{fig:retrain_freq} shows  the  impact of training horizon on the model update  frequency required  for high accuracy. For the  same driving video,  we plot  accuracy  vs.  the  model update interval ($T_{update}$) for  the  default  model  training using  $T_{horizon} = 16, 64, 256$ sec.  As expected, more  frequent model updates improves accuracy in all  cases, but the accuracy of models  trained  with  a small training horizon  ($T_{horizon} = 16$ sec) drops much more sharply as we increase $T_{update}$.

The best values of $T_{horizon}$  and $T_{update}$ depend on both the video and  the model capacity. Overall we have found that for  semantic segmentation using  mobile-friendly models, a training horizon of 3--5 minutes works well with model updates every 10--40  seconds across a variety of videos (\S\ref{sec:eval}).

\section{Adaptive Training Rate (ATR)}\label{app:atr}
\begin{figure}
\centering
\begin{tikzpicture}
\begin{axis}[xlabel=Time (sec), ylabel=Sampling Rate (fps), font=\small, height=4cm, width=\linewidth, ymin=0, xmin=-5, xmax=420,legend style={at={(0.5,1.2)}, draw=none, anchor=center}, legend columns=2]
\draw [color=red, thick] (axis cs:0.0, 0) -- (axis cs:0.0, 2);
\draw [color=red, thick] (axis cs:110.0, 0) -- (axis cs:110.0, 2);
\draw [color=red, thick] (axis cs:120.0, 0) -- (axis cs:120.0, 2);
\draw [color=red, thick] (axis cs:130.0, 0) -- (axis cs:130.0, 2);
\draw [color=red, thick] (axis cs:140.0, 0) -- (axis cs:140.0, 2);
\draw [color=red, thick] (axis cs:150.0, 0) -- (axis cs:150.0, 2);
\draw [color=red, thick] (axis cs:160.0, 0) -- (axis cs:160.0, 2);
\draw [color=red, thick] (axis cs:172.0, 0) -- (axis cs:172.0, 2);
\draw [color=red, thick] (axis cs:186.0, 0) -- (axis cs:186.0, 2);
\draw [color=red, thick] (axis cs:202.0, 0) -- (axis cs:202.0, 2);
\draw [color=red, thick] (axis cs:222.0, 0) -- (axis cs:222.0, 2);
\draw [color=red, thick] (axis cs:246.0, 0) -- (axis cs:246.0, 2);
\draw [color=red, thick] (axis cs:274.0, 0) -- (axis cs:274.0, 2);
\draw [color=red, thick] (axis cs:308.0, 0) -- (axis cs:308.0, 2);
\draw [color=red, thick] (axis cs:100.0, 0) -- (axis cs:100.0, 2);
\draw [color=red, thick] (axis cs:348.0, 0) -- (axis cs:348.0, 2);
\draw [color=black!50, dashed, thick] (axis cs:-10, 0.35) -- (axis cs:700, 0.35);
\draw [fill, color=black!50, dashed, thick] (axis cs:-10, 0.25) -- (axis cs:700, 0.25);
\node[] at (axis cs:50,0.45) {$\gamma_1$};
\node[] at (axis cs:50,0.15) {$\gamma_0$};

\addplot [black, ultra thick, mark=none, smooth] coordinates{
(0, 1.0)
(10, 1.0)
(20, 1.0)
(30, 1.0)
(40, 1.0)
(50, 1.0)
(60, 1.0)
(70, 1.0)
(80, 1.0)
(90, 1.0)
(100, 1.0)
(110, 0.8)
(120, 0.6)
(130, 0.4)
(140, 0.4)
(150, 0.3)
(152.5, 0.25)
};
\addplot [blue, ultra thick, mark=none, smooth] coordinates{
(152.5, 0.25)
(160, 0.1)
(170, 0.1)
(180, 0.3)
(190, 0.2)
(200, 0.2)
(210, 0.1)
(220, 0.3)
(230, 0.1)
(240, 0.2)
(250, 0.3)
(260, 0.1)
(270, 0.1)
(280, 0.1)
(290, 0.1)
(300, 0.1)
(310, 0.3)
(320, 0.2)
(330, 0.1)
(340, 0.2)
(350, 0.1)
(360, 0.1)
(370, 0.3)
(380, 0.2)
(390, 0.1)
(400, 0.1)
(410, 0.1)
(420, 0.1)
    };
    \legend{Normal mode, Slowdown mode}
    \end{axis}
    \end{tikzpicture}
    \caption{ATR updates of the model update intervals w.r.t. the average sampling rate over time for Vid1. Vertical lines represent model updates. Model updates become distant after entering the slowdown mode.}
    \label{fig:atr}
\end{figure}
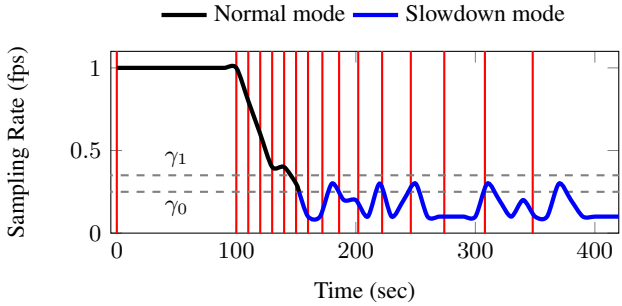

We dynamically update the model update interval $T_{update}$ for each device based on its  video characteristics.  For this purpose, for each interval $n$, we look at ASR's sampling rate decision $r_n$ (see \S\ref{sec:design_uplink}). 
A small sampling rate typically implies the scenes are changing slowly, and conversely a large sampling rate suggests fast variations.

We introduce a \emph{slowdown} mode to capture relatively stationary scenes. We enter the slowdown mode if the scenes are highly similar, $r_n < \gamma_0$, and exit this mode as variations increase, $r_n > \gamma_1$.  Our implementation uses $\gamma_0 =  0.25$~fps and  $\gamma_1 = 0.35$~fps. We start at the maximum training rate ($T_{update}=\tau_{min}$), and update the training interval every $\delta t$ seconds according to:
\begin{equation}\label{eq:train_rate_update}
    T_{update}(n+1)=\begin{cases} T_{update}(n) + \Delta, & \text{in slowdown mode} \\
    \tau_{min}, & \text{otherwise}\end{cases}
\end{equation}
This rule gradually increases $T_{update}$ by a fixed $\Delta$ (e.g., $\Delta=2$~sec)  in slowdown mode, and aggressively resets  it  to $\tau_{min}$ as soon as we exit the slowdown mode to catch up with scene changes.   

The sever communicates the newest  $T_{update}$ (and sampling rate) with the edge so that the edge device can accordingly synchronize its sample buffering and upload process (see~\S\ref{sec:design_uplink}).

In~\cref{fig:atr}, we plot the behavior of ATR algorithm for the Interview video with relatively stationary scenes from Outdoor Scenes dataset. We observe that ATR enters the slowdown mode after 150 seconds as the average sampling rate goes below the entrance threshold $\gamma_0$, and it stays in this mode as the scenes rarely change after this point. We denote the model updates using the vertical lines in this plot. ATR increases the distance between the model updates in the slowdown mode to save the training cycles for other videos. 

\section{Server Utilization}\label{app:server_utilization}
\begin{figure}
\begin{tikzpicture}
                \begin{axis}[height=4.5cm, 
                         width=1\linewidth,
                         axis lines=left,
                         ylabel near ticks,
                         xlabel near ticks,
                         xlabel={Number of clients},
                         ylabel={$\Delta$ \gls{miou} (\%)},
                         ymax=0.5,
                         grid=both,
                         legend style={at={(0,0)}, 
                           anchor=south west ,fill=white, fill opacity=0, draw opacity=1,text opacity=1, draw=none, font=\small},
                         font=\small, 
                         xmax=10,
                         ymin=-3,]
                   \input{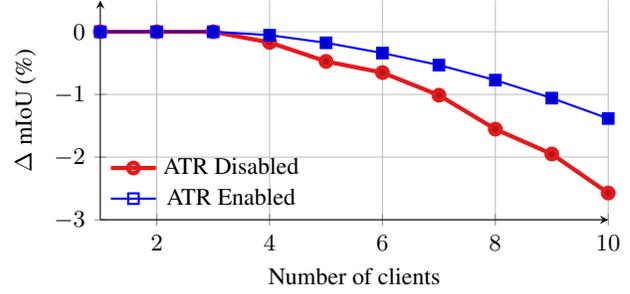}
                   \addplot+[thick, blue] coordinates{
                    (1, 0.000000)
                    (2, 0.000000)
                    (3, 0.000000)
                    (4, -0.054144)
                    (5, -0.176154)
                    (6, -0.340016)
                    (7, -0.531612)
                    (8, -0.771514)
                    (9, -1.058779)
                    (10, -1.383912)};
                    \legend{ATR Disabled, ATR Enabled};
                \end{axis}
            \end{tikzpicture}
            \caption{Average multiclient \gls{miou} degradation compared to single-client performance.}
            \label{fig:multi_client_atr}
            \end{figure}
Every user that joins a cloud server requires its own share of GPU resources for inference and training operations. GPUs are expensive. At the current time, renting a GPU like the NVIDIA Tesla V100 in the cloud costs at least \$1 per hour. Hence, it is important to use server GPU resources efficiently and serve multiple edge devices per GPU to keep per-user cost low. 

In our prototype, we use a simple strategy that iterates in a round-robin fashion across multiple video sessions, completing one inference and training step per session. By allowing only one process to access the GPU at a time, we minimize context switching overhead. In~\cref{fig:multi_client_atr} we show the decrease (w.r.t. single client) in average \gls{miou} when different number of clients share a GPU. We observe that even with a simple round-robin scheduling algorithm, \sys scales to up to 7 edge devices on a single V100 GPU with less than 1\% loss in \gls{miou} without adaptive training rate (ATR) enabled. Enabling ATR (see Appendix~\ref{app:atr}) increases the number of supported edge devices to 9. Note that these results depend on the distribution of the videos and for this purpose, we have assumed a uniform sampling of videos from the Outdoor Scenes dataset and reported the average result of multiple runs here. As most of the videos in this dataset (5 out of 7) tend to experience relatively high levels of scene dynamic, majority of videos get high training frequency. Hence, we expect to be able to support at least equal or even more devices by prioritizing certain videos that need more frequent model updates over the stationary ones in real-world distribution of videos.  

Furthermore, we note that ASR (see~\S\ref{sec:design_uplink}) also significantly reduces the overhead of running teacher inference over redundant frames at the server. The impact is particularly pronounced because the teacher model usually runs at high input resolution and consumes a significant amount of GPU time (up to 200 ms for labeling each frame on an NVIDIA V100 GPU for the task of semantic segmentation).

\section{Uplink Sampling Rate}\label{app:sampling_rate_cont} 
\begin{figure}[!t]
    \centering
    \begin{tikzpicture}
        \begin{axis}[
        font=\small, 
        height=4cm, width=\linewidth, ylabel=\textbf{CDF}, xlabel=\textbf{Avg. Sampling Rate (fps)}, xlabel near ticks, ylabel near ticks, xmax=1.1, xmin=0]
        \addplot[violet, ultra thick, mark=none] coordinates{
        (0, 0)
(0.23081503333333334, 0.02564102564102564)
(0.2568891, 0.05128205128205128)
(0.28296316666666665, 0.07692307692307693)
(0.3090372333333333, 0.07692307692307693)
(0.3351113, 0.15384615384615385)
(0.3611853666666667, 0.1794871794871795)
(0.3872594333333333, 0.20512820512820512)
(0.41333349999999996, 0.20512820512820512)
(0.43940756666666664, 0.20512820512820512)
(0.4654816333333333, 0.20512820512820512)
(0.4915557, 0.20512820512820512)
(0.5176297666666667, 0.20512820512820512)
(0.5437038333333333, 0.23076923076923078)
(0.5697779000000001, 0.23076923076923078)
(0.5958519666666666, 0.23076923076923078)
(0.6219260333333334, 0.23076923076923078)
(0.6480001, 0.23076923076923078)
(0.6740741666666668, 0.2564102564102564)
(0.7001482333333333, 0.28205128205128205)
(0.7262223000000001, 0.3076923076923077)
(0.7522963666666667, 0.3333333333333333)
(0.7783704333333334, 0.3333333333333333)
(0.8044445, 0.46153846153846156)
(0.8305185666666667, 0.46153846153846156)
(0.8565926333333334, 0.46153846153846156)
(0.8826666999999999, 0.5128205128205128)
(0.9087407666666667, 0.5641025641025641)
(0.9348148333333333, 0.6153846153846154)
(0.9608889, 0.6410256410256411)
(0.9869629666666666, 1.0)
(2,1)
};
        \end{axis}
    \end{tikzpicture}
    \caption{Cumulative distribution of average ASR sampling rate across all videos.}
    \label{fig:sampling_rate_cdf}
\end{figure}
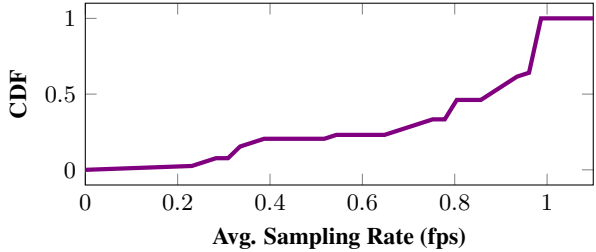

\Cref{fig:sampling_rate_cdf} shows the distribution of ASR's average sampling rate across different videos in four datasets. Notice that we set the ASR's maximum sampling rate (see~\S\ref{sec:design_uplink}), $r_{max}$, to 1 fps as our results show sampling faster than 1 frame-per-second provides negligible improvement in accuracy along increasing bandwidth usage and server inference overhead. We use $r_{min}=0.1$ fps.

\end{appendices}
\end{document}